\documentclass[a4paper,fleqn]{cas-dc}

\usepackage[authoryear]{natbib}
\usepackage[colorinlistoftodos]{todonotes}
\usepackage{xcolor}
\usepackage{color}
\usepackage{soul}
\usepackage{graphicx}
\usepackage{float}
\usepackage{amsmath}
\usepackage{amsfonts}
\usepackage{amssymb}
\usepackage{graphicx}
\usepackage{subcaption}
\usepackage{float}
\usepackage{algorithm}
\usepackage{algorithmic}

\def\tsc#1{\csdef{#1}{\textsc{\lowercase{#1}}\xspace}}
\tsc{WGM}
\tsc{QE}

\begin{document}
\let\WriteBookmarks\relax
\def\floatpagepagefraction{1}
\def\textpagefraction{.001}

\shorttitle{TFN-PV-Temporal Fusion–NBEATS Photovoltaic Forecasting Model}    

\shortauthors{}  

\title [mode = title]{A Hybrid TFT-NBEATS Framework for Robust Multi-Horizon Photovoltaic Power Forecasting}  


\tnotetext[1]{**These authors contribute equally} 
%
\author[1]{Shafi Bin Sultan**}
\ead{shafibinsultan0207@gmail.com}
\author[2]{Sabik Bin Sultan**}
\ead{sabikbinsultan@yahoo.com}

\author[3]{Mohammad Tajrian Arafat Alvi}
\ead{arafat3568@gmail.com}

\author[4]{Rayhanul Islam Sony}
\ead{rsony23@my.trine.edu}

\author[5]{Abdulah Al Numan}
\ead{Numan290.AaN@gmail.com}
\cormark[1]

\affiliation[1]{organization={St. Joseph Higher Secondary School},
            city={Dhaka},
            country={Bangladesh}}

\affiliation[2]{organization={Bangladesh Air Force Shaheen College, Kurmitolla},
            city={Dhaka},
            country={Bangladesh}}

\affiliation[3]{organization={Computer Science \& Engineering, Jahangirnagar University},
            city={Savar},
            country={Bangladesh}}

\affiliation[4]{organization={Trine University},
            city={Angola},
            state={Indiana},
            country={USA}}

\affiliation[5]{organization={Sembcorp North-West Power Company Ltd},
            city={Dhaka},
            country={Bangladesh}}

\cortext[1]{Corresponding author}

\begin{abstract}
Accurate multi-horizon photovoltaic (PV) power forecasting is essential for grid stability, energy scheduling, and the efficient operation of PV–battery–grid systems. However, existing models often fail to jointly capture short-term variability and long-term temporal patterns, while lacking unified multi-horizon capability and interpretability. To address these limitations, this study proposes a novel hybrid deep learning framework that integrates a Temporal Fusion Transformer (TFT) and N-BEATS within an attention-based fusion architecture. The proposed model leverages TFT to learn short-term temporal dependencies and dynamic feature importance, while N-BEATS captures long-term trend and seasonal components through interpretable basis decomposition. An adaptive attention mechanism dynamically fuses both branches, enabling balanced multi-scale learning. The framework simultaneously generates forecasts for 30-minute, 2-hour, 6-hour, 12-hour, and 24-hour horizons within a single unified pipeline. The model is evaluated using multi-year real-world datasets from the Desert Knowledge Australia Solar Centre (DKASC). Experimental results demonstrate superior performance over statistical, machine learning, and deep learning baselines. The proposed model achieves RMSE values of 42.35, 46.27, 52.29, and 48.11 across the 2016–2017, 2023, 2024, and 2025–2026 datasets, respectively, corresponding to a 35–40\% reduction compared to the best baseline (TFT). Additionally, MAPE remains below 5\% across all datasets, indicating high predictive accuracy and stability. The model also provides interpretability through attention visualization and trend–seasonality decomposition, and is deployed on a cloud-based platform for real-time forecasting. These results demonstrate that the proposed framework is robust, interpretable, and suitable for practical renewable energy applications.

\end{abstract}


\maketitle

\section{Introduction}
 
Solar photovoltaic (PV) technology is a leading driver of the global energy transition. In 2024, solar PV added a record 451.9~GW of new capacity, raising cumulative global installations to 1,865~GW \cite{IRENA2025}. Solar PV and wind together are projected to supply 95\% of new renewable capacity through 2030, supported by generation costs now below most fossil alternatives \cite{IEA2024}. Despite this growth, large-scale PV integration introduces serious operational challenges. PV output is intermittent, driven by solar irradiance, cloud
dynamics, and ambient temperature \cite{Nadeem2024}. These meteorological
variables vary rapidly and non-linearly, causing abrupt fluctuations in power
supply \cite{Du2024}. Such instability threatens frequency regulation, power
balancing, and economic dispatch in modern grids with high renewable penetration
\cite{Nadeem2024, Rizk2024}.
 
Accurate PV power forecasting is essential for managing this variability. Forecasts spanning minutes to 24~hours ahead support grid balancing, unit commitment, battery storage scheduling, and electricity market operations
\cite{Alorf2025, Mbey2024}. In hybrid PV-battery-grid systems, predictions
across multiple timescales are needed simultaneously to coordinate battery
charging, peak saving, and grid dispatch \cite{Lim2021}. Forecast errors at any
horizon propagate into costly reserve activations and unnecessary curtailment
\cite{Salman2024}.
 
Recent research categorizes photovoltaic forecasting methods into three main groups: physical, statistical, and data-driven approaches \cite{Du2024, Asiedu2024}. Physical models apply solar geometry and atmospheric equations but rely heavily on NWP quality and
site-specific parameters \cite{Du2024}. Statistical methods such as ARIMA
perform well over short horizons but struggle with the non-stationary, nonlinear
behaviour of PV generation \cite{Park2023}. Machine learning (ML) and deep
learning (DL) models have therefore become the dominant paradigm, learning
complex mappings directly from historical meteorological and power data
\cite{Nadeem2024, Du2024}.
 
Recent studies have advanced towards hybrid architectures that combine multiple
complementary model components. \citet{Asiedu2024} showed that hybrid
XGBoost-Random~Forest stacking outperformed all individual models across
day-ahead, week-ahead, and month-ahead horizons on a 180~kWp PV system.
\citet{Mondal2024} coupled XGBoost with LSTM under a physics-informed pipeline,
incorporating irradiance-aware resampling and domain-guided imputation.
Evaluated on the UNISOLAR dataset of 2.7~million 15-minute samples from 42
sites, this framework achieved an RMSE of 2.57~kWh ($R^{2} = 0.934$), a
17.84\,\% gain over standalone LSTM \cite{Mondal2024}. \citet{Guermoui2024}
fused five decomposition algorithms with a deep CNN across three PV plants,
showing that multi-frequency decomposition consistently outperforms any
single-method baseline. \citet{Alorf2025} combined Variational Mode
Decomposition (VMD) with a modified Temporal Fusion Transformer (TFT) for
improved multi-horizon accuracy under variable weather. \citet{Nadeem2024b}
addressed spatial NWP bias through an attention-enhanced CNN--LSTM with
XGBoost-based downscaling, achieving consistent RMSE and MAE improvements
across multiple stations. For interpretability, \citet{Rizk2024} augmented LSTM with LIME-based explainable AI, showing model transparency is achievable without
accuracy loss. \citet{Oreshkin2020} established with N-BEATS that interpretable
trend-seasonality decomposition via basis expansion can match black-box DL
models on multi-horizon benchmarks. \citet{Lim2021} subsequently introduced
TFT, combining recurrent encoding with multi-head self-attention and
variable-selection gating for interpretable multi-horizon forecasting.
 
Despite these advances, three critical limitations persist. First,
most studies treat each forecast horizon as an independent problem, deploying
separate models for intra-hour, hour-ahead, and day-ahead prediction
\cite{DL_review2024}. Real-world PV-battery-grid systems require coherent predictions across multiple timescales simultaneously \cite{Lim2021,Alorf2025}. Second, PV generation exhibits dynamics at two distinct
temporal scales including short-term fluctuations from cloud movement and irradiance
variation, and long-term trends from daily solar cycles and seasonal patterns.
Models optimized for one scale routinely degrade on the other, resulting in
temporal-scale bias \cite{Nadeem2024, Asiedu2024, Park2023}. In last,
high-accuracy DL models commonly operate as black boxes. Their opaque decision
processes undermine operator trust and limit deployment in real energy management
systems \cite{DL_review2024, Rizk2024}. Existing studies either rely on decomposition-heavy pipelines, which increase computational complexity, or focus on single-scale learning, limiting adaptability in real-world multi-horizon scenarios.

No existing framework simultaneously resolves all three limitations. Current
hybrid models are predominantly single-horizon, fail to balance learning across
temporal scales, and lack interpretable outputs. This study addresses these gaps
within a single unified, deployment-ready pipeline. By addressing these limitations, this study pursues four objectives. \textbf{(O1)}~Develop a unified framework
that simultaneously forecasts PV power at 30-minute, 2-hour, 6-hour, 12-hour,
and 24-hour horizons. \textbf{(O2)}~Design a dual-branch architecture combining
TFT for short-term dependency modelling with N-BEATS for trend-seasonality
decomposition to achieve balanced multi-scale learning. \textbf{(O3)}~Incorporate
attention visualisation and basis-function decomposition for partial model
interpretability. \textbf{(O4)}~Validate robustness across multiple real-world
PV datasets spanning different years and climatic conditions. Unlike existing hybrid models, the proposed framework introduces a joint multi-horizon learning mechanism with dynamic attention-based fusion, enabling simultaneous optimization across temporal scales within a single unified architecture.
 
The main contributions of this study are as follows:
 
\begin{itemize}
 
  \item \textbf{Unified multi-horizon framework.}
        A single pipeline simultaneously generates PV power forecasts at five
        operational horizons (30~min, 2~h, 6~h, 12~h, and 24~h), eliminating
        the need for separate horizon-specific models.
 
  \item \textbf{Hybrid dual-branch architecture (TFT + N-BEATS).}
        TFT handles short-term temporal dependencies via recurrent encoding and
        multi-head attention \cite{Lim2021}. N-BEATS provides interpretable
        trend--seasonality decomposition through polynomial and Fourier basis
        blocks \cite{Oreshkin2020}.
 
  \item \textbf{Attention-based fusion mechanism.}
        A trainable fusion layer dynamically weights the contributions of both
        branches, enabling adaptive learning across temporal scales without
        manual tuning.
 
  \item \textbf{Physics-aware preprocessing.}
        Irradiance-guided resampling and operational-status-based imputation
        reduce sensitivity to missing values and irregular sampling in real-world
        PV data \cite{Mondal2024}.
 
  \item \textbf{Partial model interpretability.}
        Attention weight visualisation and N-BEATS decomposition outputs expose
        temporal importance and forecast components, supporting operator trust
        and system diagnostics.
 
  \item \textbf{Multi-dataset validation.}
        The framework is evaluated across multiple real-world PV datasets
        spanning different years, sites, and climatic conditions.
 
  \item \textbf{Deployment-ready implementation.}
        The model is deployed via Hugging Face infrastructure for practical
        real-time use in hybrid PV-battery-grid systems.
 
\end{itemize}
 
The remainder of this paper is organised as follows. Section \ref{sec2} reviews related
work. Section~3 describes datasets and preprocessing. Section~4 presents the
proposed architecture. Section~5 details experimental setup and metrics.
Section~6 reports results and interpretability analysis. Section~7 discusses
deployment implications, and Section~8 concludes the study.

\section{Related Work} \label{sec2}

Photovoltaic (PV) power generation has gained significant research attention due to its important role in reducing carbon emissions and supporting the transition toward sustainable energy systems. However, the intermittent and weather-dependent nature of solar energy introduces substantial uncertainty in power generation, making accurate PV power forecasting essential for reliable grid operation, energy scheduling, and power system planning. Over the past decade, numerous forecasting approaches have been proposed, ranging from traditional statistical models to advanced machine learning and deep learning architectures. More recently, research has also explored multi-horizon forecasting strategies and explainable artificial intelligence (XAI) techniques to improve both prediction accuracy and model transparency.

\subsection{Machine Learning Based PV Forecasting}

Machine learning models have been widely applied to photovoltaic (PV) power forecasting due to their ability to capture nonlinear relationships between meteorological variables and PV generation. Algorithms such as Support Vector Regression (SVR), Random Forest, Gradient Boosting, and K-Nearest Neighbors (KNN) have demonstrated improved predictive performance compared with traditional statistical models. For example, Kim et al. \cite{kim2026evaluation} proposed a forecasting framework combining Model Output Statistics (MOS) with a Random Forest model to correct systematic errors in meteorological forecasts, thereby improving the accuracy of PV power prediction. Similarly, Salehin et al. \cite{salehin2026comparative} developed a hybrid KNN–SVM model in which KNN identifies similar historical patterns while SVM performs local regression to estimate solar power generation. In addition, ensemble learning methods such as Gradient Boosting and Random Forest have demonstrated strong performance in PV forecasting due to their ability to handle nonlinear relationships and noisy meteorological data.
Although machine learning approaches significantly improve forecasting performance by modeling nonlinear feature interactions, these methods often rely heavily on manual feature engineering and typically struggle to capture long-term temporal dependencies present in photovoltaic time-series data. Moreover, their ability to represent complex temporal dynamics remains limited when forecasting over multiple future time horizons, which restricts their applicability in real-world energy management scenarios.

\subsection{Deep Learning Approaches for Photovoltaic Power Forecasting}

To overcome the limitations of conventional machine learning approaches in modeling sequential data, deep learning architectures have increasingly been adopted for photovoltaic forecasting due to their ability to automatically learn hierarchical feature representations and temporal dependencies from time-series data.

Recent studies have particularly focused on hybrid deep learning frameworks that combine multiple neural architectures to enhance forecasting performance. For example, Yousuf et al. \cite{yousuf2026hybrid} introduced a hybrid XGBoost–LSTM architecture in which XGBoost models nonlinear feature interactions while LSTM captures long-term temporal dependencies in photovoltaic power generation. Other hybrid architectures have also been explored in the literature. Kim and Hau \cite{kim2026advancing} developed a CNN–DNN model for PV power prediction and applied SHAP-based explainability to analyze feature importance. Baran \cite{baran2026deep} proposed a CNN–LSTM model for detecting PV system performance degradation using multi-sensor measurements. Fennane et al. \cite{fennane2026improving} introduced a hybrid LSTM–TCN architecture capable of modeling both short-term and long-term temporal patterns, while Oumiguil et al. \cite{oumiguil2026short} proposed a CNN–GRU–XGBoost framework for short-term PV power forecasting.

Despite their improved predictive performance, many hybrid deep learning models remain limited to specific datasets or single forecasting horizons. Furthermore, most architectures are optimized for either short-term or day-ahead forecasting tasks, making it challenging to simultaneously capture both rapid short-term fluctuations and broader temporal patterns in photovoltaic generation.

\subsection{Multi-Horizon Photovoltaic Power Forecasting}

While deep learning architectures have significantly improved photovoltaic power forecasting, many existing studies still focus primarily on single-step or single-horizon prediction tasks. However, real-world energy management systems require forecasts across multiple future time horizons to support grid stability, battery scheduling, and operational decision-making.

To address this challenge, recent studies have explored multi-horizon forecasting frameworks. For instance, Cisse et al. \cite{cisse2026optimized} proposed a hybrid TCN–GRU–Random Forest framework for short-term multi-horizon PV forecasting using operational solar plant data. Han et al. \cite{han2025enhanced} introduced a decomposition-based forecasting approach integrating ICEEMDAN with hierarchical frequency neural networks to capture non-stationary temporal characteristics in PV generation. Similarly, Khan et al. \cite{khan2025interpretable} developed an interpretable hybrid architecture combining CNN, LSTM, and Transformer models with SHAP-based explainability for multi-horizon renewable energy forecasting. In a broader energy forecasting context, Gawusu and Zhang \cite{gawusu2026transformer} demonstrated that transformer-based architectures can effectively model long-range temporal dependencies in multi-horizon power generation prediction.
Although these studies highlight the potential of multi-horizon forecasting, many existing approaches rely on complex decomposition or ensemble strategies that increase model complexity and computational overhead. Additionally, achieving balanced learning across multiple temporal scales remains a challenging problem, particularly when models must simultaneously capture short-term fluctuations and longer-term generation trends.

\subsection{Explainable Artificial Intelligence in PV Forecasting}

In addition to forecasting accuracy, model interpretability has become an important requirement in photovoltaic power forecasting, particularly for operational energy management systems where understanding the influence of environmental variables is essential for decision-making.

Rizk-Allah et al. \cite{rizk2024explainable} proposed an explainable photovoltaic forecasting framework that integrates Long Short-Term Memory (LSTM) networks with an Equilibrium Optimizer (EO) for hyperparameter tuning and Local Interpretable Model-Agnostic Explanations (LIME) for model interpretability. The LSTM model captures temporal dependencies in solar power generation, while the LIME-based explanation method identifies key environmental variables influencing prediction outcomes. Similarly, Alagarsamy et al. \cite{alagarsamy2025xai} introduced an explainable photovoltaic energy management framework for smart city applications. Their model-chain architecture integrates recurrent neural networks (RNN) for long-term forecasting of global horizontal irradiance (GHI), followed by regression-based modeling to estimate photovoltaic power generation. The framework employs LIME explanations to improve model transparency and achieved improved forecasting performance with RMSE and MAE values of 91.40 and 47.32 respectively. More recently, Bakht et al. \cite{bakht2025advanced} proposed an automated machine learning framework that integrates Auto-ML with SHAP-based explainable artificial intelligence for photovoltaic power prediction. Their study used real PV plant data with environmental parameters and reported that the Extra Tree model achieved the highest prediction accuracy of 91\%, outperforming Random Forest (85\%), XGBoost (75\%), and Decision Tree (68\%). The SHAP analysis further provided interpretable insights into the influence of meteorological variables on PV power generation.

Although these approaches improve model transparency, most explainability techniques are applied as post-hoc analysis methods that interpret predictions after the forecasting process. As a result, they provide limited insights into the internal temporal learning behavior of forecasting architectures, highlighting the need for models that combine strong predictive capability with more integrated interpretability mechanisms.

\subsection{Research Gap}

Despite significant progress in photovoltaic power forecasting research, several important challenges remain. As summarized in \autoref{tab:literature_summary}, many existing studies address forecasting tasks at a single temporal horizon, which limits their practical usefulness for real-world PV–battery–grid energy systems that require predictions across multiple future time scales. Furthermore, current deep learning architectures often struggle to achieve balanced learning between short-term fluctuations caused by rapid irradiance changes and longer-term temporal patterns driven by daily and seasonal cycles. Therefore, developing a robust forecasting framework that can simultaneously learn multi-scale temporal patterns while maintaining interpretability remains an important research challenge in photovoltaic power forecasting.

To address these limitations, this study proposes a unified multi-horizon photovoltaic power forecasting framework that integrates complementary temporal learning capabilities within a hybrid architecture. The proposed model is designed to simultaneously capture short-term variability and long-term temporal dependencies while providing interpretable insights into the forecasting process, enabling more reliable and scalable PV power prediction for practical energy management systems.

\begin{table*}[t]
\centering
\caption{Summary of representative studies in photovoltaic power forecasting}
\small
\setlength{\tabcolsep}{4pt}
\renewcommand{\arraystretch}{1.05}
\begin{tabular}{p{2.8cm} p{3.2cm} p{2cm} p{3cm} p{4cm}}
\hline
\textbf{Study} & \textbf{Method} & \textbf{Forecast Horizon} & \textbf{Dataset} & \textbf{Key Limitation} \\
\hline

Kim et al. \cite{kim2026evaluation} & MOS + Random Forest & Short-term & Rooftop PV (Korea) & Limited temporal modeling \\

Salehin et al. \cite{salehin2026comparative} & KNN–SVM & Short-term & NASA POWER + local data & Requires manual feature engineering \\

Yousuf et al. \cite{yousuf2026hybrid} & XGBoost + LSTM & Multi-step & UNISOLAR (42 PV sites) & Strong dependence on lag features \\

Fennane et al. \cite{fennane2026improving} & LSTM + TCN & Short-term & PVGIS (Morocco) & Focus on short-term forecasting \\

Cisse et al. \cite{cisse2026optimized} & TCN + GRU + RF & Multi-horizon & Solar plant (Australia) & Ensemble complexity \\

Han et al. \cite{han2025enhanced} & ICEEMDAN + HFNN & Multi-horizon & DKASC, Solar I & Requires signal decomposition \\

Khan et al. \cite{khan2025interpretable} & CNN + LSTM + Transformer + SHAP & Multi-horizon & Multi-source datasets & High computational cost \\

Bakht et al. \cite{bakht2025advanced} & AutoML + SHAP & Single-step & PV plant (Malaysia) & Limited multi-horizon capability \\

Rizk-Allah et al. \cite{rizk2024explainable} & LSTM + EO + LIME & Short-term & PV environmental data & Post-hoc interpretability \\

\hline

\end{tabular}
\label{tab:literature_summary}
\end{table*}
\section{Methodology}
\begin{figure*}
    \centering
    \includegraphics[width=\linewidth]{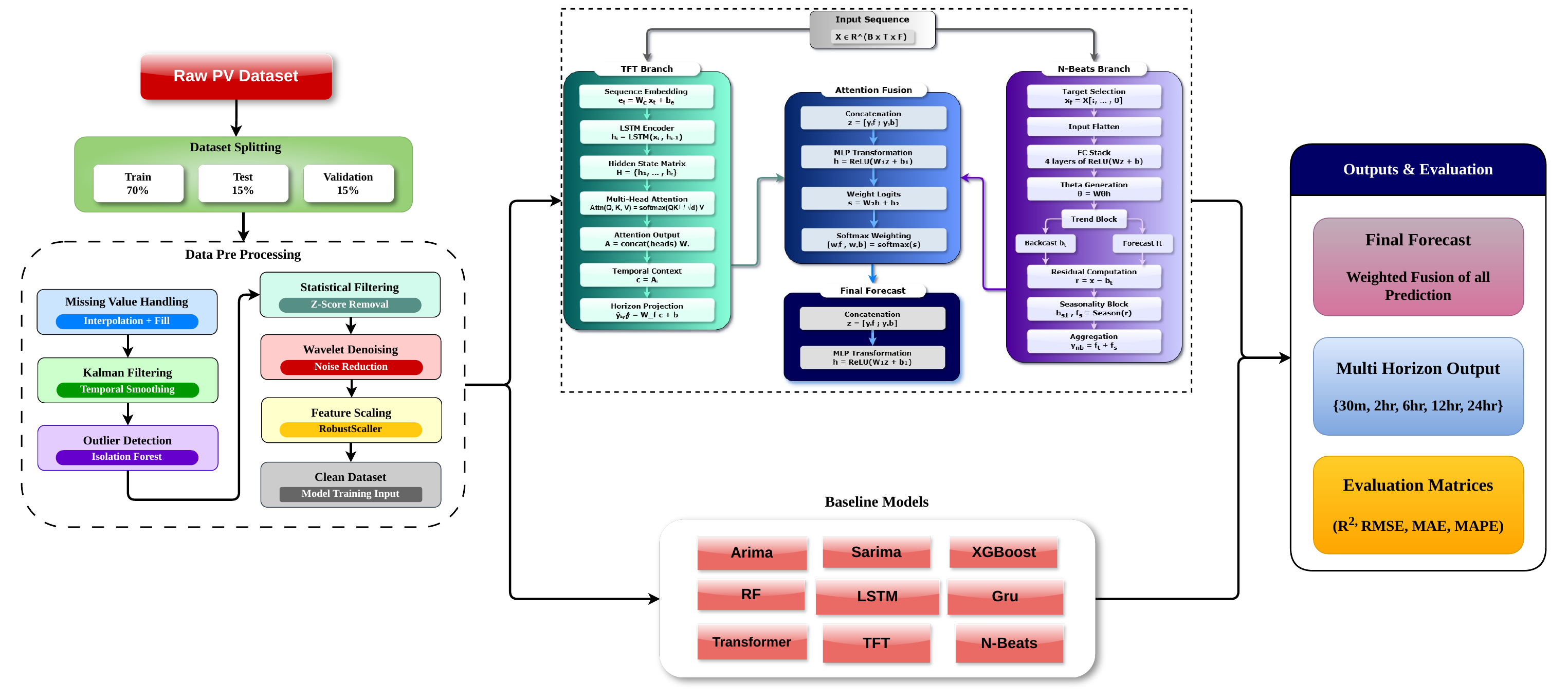}
    \caption{Proposed hybrid multi-horizon photovoltaic (PV) forecasting framework. The pipeline includes data preprocessing, dual-branch modeling using TFT and N-BEATS, attention-based fusion, and multi-horizon prediction with evaluation against baseline models.}
    \label{fig:workflow}
\end{figure*}

This study proposes a unified hybrid deep learning framework for multi-horizon photovoltaic (PV) power forecasting, integrating data preprocessing, dual-branch temporal modeling, and attention-based fusion within a single end-to-end pipeline, as illustrated in \autoref{fig:workflow}. The workflow begins with raw PV and meteorological data, which are chronologically split into training, validation, and testing subsets to preserve temporal integrity. A comprehensive preprocessing pipeline is then applied, including missing value imputation, Kalman filtering for temporal smoothing, anomaly detection using Isolation Forest, statistical outlier removal, wavelet-based denoising, and robust feature scaling, ensuring high-quality input for model training. The cleaned time-series data are fed into a hybrid architecture consisting of two complementary branches: a Temporal Fusion Transformer (TFT) branch for modeling short-term temporal dependencies and dynamic feature importance, and an N-BEATS branch for capturing long-term trend and seasonal components through interpretable decomposition. The outputs of both branches are combined using an attention-based fusion mechanism, which adaptively weights their contributions based on the temporal context. The final model generates simultaneous forecasts across multiple horizons (30 minutes to 24 hours) within a unified framework. To validate performance, the proposed approach is benchmarked against a range of baseline models, including statistical, machine learning, and deep learning methods, and evaluated using standard regression metrics.

\subsection{Dataset Description}
The photovoltaic dataset used in this study was obtained from the Desert Knowledge Australia Solar Centre (DKASC) [1], located in Alice Springs, Northern Territory, Australia. The DKASC facility is a well-known solar technology demonstration site that hosts multiple photovoltaic (PV) installations operating under the arid climatic conditions of central Australia. The data are publicly available through the DKASC data portal and include both PV operational measurements and associated meteorological observations. The dataset is recorded at 5-minute intervals, enabling high temporal resolution of photovoltaic system performance and environmental conditions. This temporal resolution allows detailed observation of short-term variations in solar power output caused by changing atmospheric conditions and solar irradiance. The dataset contains several variables describing both the PV system's electrical generation and the surrounding weather conditions. The primary variable is active power output, which represents the actual electrical power generated by the photovoltaic system. In addition, the dataset includes important meteorological variables such as global horizontal irradiance (GHI), diffuse horizontal irradiance (DHI), ambient temperature, relative humidity, and wind speed, which are known to influence photovoltaic power generation. To evaluate the robustness and generalization capability of the proposed forecasting framework, experiments are conducted using multiple datasets collected across different years, as summarized in \autoref{tab:dataset_time_coverage}. Specifically, photovoltaic generation data from 2016–2017, 2023, 2024, and 2025–2026 are analyzed independently. This multi-year dataset configuration allows the study to capture seasonal variations, long-term temporal dynamics, and year-to-year variability in photovoltaic power generation. Consequently, evaluating the forecasting model across different datasets provides a more comprehensive assessment of its robustness and practical applicability under varying environmental conditions.

\begin{table}[h]
\centering
\caption{Time coverage of the multi-year photovoltaic datasets used in this study.}
\label{tab:dataset_time_coverage}
\begin{tabular}{ll}
\toprule
\textbf{Dataset} & \textbf{Time Coverage} \\
\midrule
DKASC-2016-17 & 2016--2017 \\
DKASC-2023    & 2023 \\
DKASC-2024    & 2024 \\
DKASC-2025-26 & 2025--2026 \\
\bottomrule
\end{tabular}
\end{table}

\begin{table*}
\centering
\caption{Key photovoltaic and meteorological variables used in the forecasting model.}
\label{tab:pv_features}
\begin{tabular}{lp{9cm}}
\toprule
\textbf{Feature} & \textbf{Description} \\
\midrule
Active Power & Electrical power generated by the PV system (target variable) \\
Global Horizontal Irradiance & Solar radiation received on a horizontal surface \\
Diffuse Horizontal Irradiance & Scattered solar radiation component \\
Ambient Temperature & Air temperature measured at the weather station \\
Relative Humidity & Percentage of atmospheric moisture \\
Wind Speed & Wind speed measured at the weather station \\
\bottomrule
\end{tabular}
\end{table*}

The main variables used in this study are summarized in \autoref{tab:pv_features}, which include photovoltaic power generation and key meteorological parameters that directly influence solar energy production. Among these variables, active power output is used as the target variable for forecasting, while the remaining environmental features provide contextual information about atmospheric and solar conditions. Based on these inputs, the forecasting task is formulated as a multi-horizon prediction problem, where the model generates predictions for multiple future time intervals, as presented in \autoref{forecasting}. The selected horizons range from 30 minutes to 24 hours, enabling the model to capture both short-term fluctuations and longer-term temporal patterns in photovoltaic power generation.

\begin{table*}[t]
\centering
\caption{Forecasting horizons, steps, and purposes in photovoltaic power forecasting}
\label{forecasting}
\small
\setlength{\tabcolsep}{8pt} 
\renewcommand{\arraystretch}{1.2} 
\begin{tabular}{p{3cm} p{3cm} p{7cm}}
\hline
\textbf{Horizon} & \textbf{Forecast Steps} & \textbf{Purpose} \\
\hline
30 Minutes & 6 steps & Real-time grid stability \\
2 Hours & 24 steps & Short-term dispatching \\
6 Hours & 72 steps & Intraday planning \\
12 Hours & 144 steps & Day-ahead preparation \\
24 Hours & 288 steps & Long-term storage management \\
\hline
\end{tabular}
\end{table*}

\subsection{Dataset Splitting}
To ensure reliable model training and unbiased evaluation, each photovoltaic dataset was partitioned using a structured chronological data splitting strategy. As summarized in \autoref{tab:data_split}, each dataset was divided chronologically into 70\% training, 15\% validation, and 15\% testing subsets, while preserving the temporal order of the observations. This chronological splitting approach is essential for time-series forecasting tasks, as it prevents information leakage from future data into the training phase. The training set is used for model learning, the validation set supports hyperparameter tuning and model selection, and the test set provides an unbiased evaluation of forecasting performance. The corresponding sample counts for each subset are reported in \autoref{tab:data_split}, reflecting the different time spans covered by each dataset. Overall, this splitting configuration ensures a balanced distribution of samples while maintaining the temporal integrity of the photovoltaic power generation data.

\begin{table*}
\centering
\caption{Chronological data splitting and sample sizes for each photovoltaic dataset.}
\label{tab:data_split}
\begin{tabular}{lccc}
\toprule
\textbf{Dataset} & \textbf{Train Samples (70\%)} & \textbf{Validation Samples (15\%)} & \textbf{Test Samples (15\%)} \\
\midrule
DKASC-2016-17 & 147743 & 31660 & 31659 \\
DKASC-2023    & 71331  & 15286 & 15285 \\
DKASC-2024    & 71331  & 15286 & 15285 \\
DKASC-2025-26 & 73770  & 15808 & 15808 \\
\bottomrule
\end{tabular}
\end{table*}

\subsection{Data Preprocessing}
\begin{figure*}
    \centering
    \includegraphics[width=0.7\linewidth]{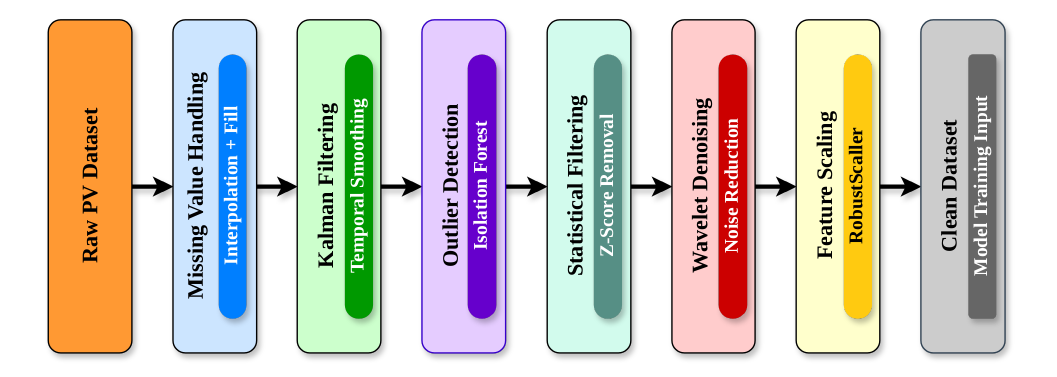}
    \caption{Data preprocessing pipeline applied to the photovoltaic dataset for PV power forecasting.}
    \label{fig:preprocessing_pipeline}
\end{figure*}

High-resolution photovoltaic (PV) measurements often contain missing observations, measurement noise, and abnormal spikes caused by sensor malfunctions, communication interruptions, or environmental disturbances. To improve data quality and ensure reliable model training, a structured preprocessing pipeline was applied to the raw PV dataset. The overall workflow of the preprocessing stages is illustrated in \autoref{fig:preprocessing_pipeline}. All numerical variables used as model inputs were processed through this preprocessing pipeline before model training.

Key preprocessing steps are summarized as follows:
\subsubsection{Missing Value Handling}

Missing observations commonly occur in PV datasets due to sensor faults, communication failures, or maintenance activities. To restore the continuity of the time series, missing values were estimated using linear interpolation, which approximates intermediate values between neighboring observations. Remaining gaps were filled using forward and backward propagation to preserve temporal consistency. This step ensures that the time-series structure remains intact before further processing.

\subsubsection{Kalman Filtering (Temporal Smoothing)}

After reconstructing missing observations, a Kalman filtering approach was applied to smooth the time-series data and reduce measurement noise. The Kalman filter estimates the hidden system state by recursively updating predictions based on observed measurements:

\begin{equation}
\hat{x}_t = A\hat{x}_{t-1} + K_t (y_t - H\hat{x}_{t-1})
\end{equation}

where $\hat{x}_t$ represents the estimated state at time $t$, $y_t$ denotes the observed measurement, $A$ is the state transition matrix, and $K_t$ is the Kalman gain. This method improves temporal consistency and removes short-term measurement fluctuations.

\subsubsection{Outlier Detection (Isolation Forest)}

PV datasets may contain abnormal spikes or sensor anomalies that can distort model learning. To detect such irregular observations, an Isolation Forest algorithm was applied. This unsupervised anomaly detection method isolates abnormal samples by recursively partitioning the data space. Observations identified as anomalies were removed from the dataset to prevent the forecasting model from learning misleading patterns.

\subsubsection{Statistical Outlier Filtering (Z-score Analysis)}

In addition to machine learning–based anomaly detection, a statistical filtering step was applied using z-score analysis to remove extreme values. The z-score is defined as:

\begin{equation}
z = \frac{x - \mu}{\sigma}
\end{equation}

where $x$ is the observation, $\mu$ is the mean, and $\sigma$ is the standard deviation. Observations with $|z|$ greater than a predefined threshold were treated as extreme values and removed to improve data reliability.

\subsubsection{Wavelet Denoising}

High-frequency solar measurements often contain noise caused by sensor sensitivity and atmospheric variability. To address this issue, wavelet-based denoising was applied. Using the Discrete Wavelet Transform (DWT), the signal was decomposed into multiple frequency components:

\begin{equation}
x(t) = \sum_k a_{j,k}\phi_{j,k}(t) + \sum_j \sum_k d_{j,k}\psi_{j,k}(t)
\end{equation}

where $a_{j,k}$ and $d_{j,k}$ represent approximation and detail coefficients. Noise-dominated coefficients were suppressed using threshold filtering, and the signal was reconstructed to produce a smoother time series while preserving meaningful temporal patterns.

\subsubsection{Feature Scaling (RobustScaler)}

Since PV power, irradiance, temperature, and wind variables have different numerical ranges, feature normalization was applied using RobustScaler. This scaling method uses the median and interquartile range (IQR) rather than mean and standard deviation:

\begin{equation}
x' = \frac{x - \text{median}(x)}{\text{IQR}(x)}
\end{equation}

This approach reduces the influence of outliers and improves numerical stability during model training.

\subsection{Proposed Forecasting Framework}
\begin{figure*}
    \centering
    \includegraphics[width=0.7\linewidth]{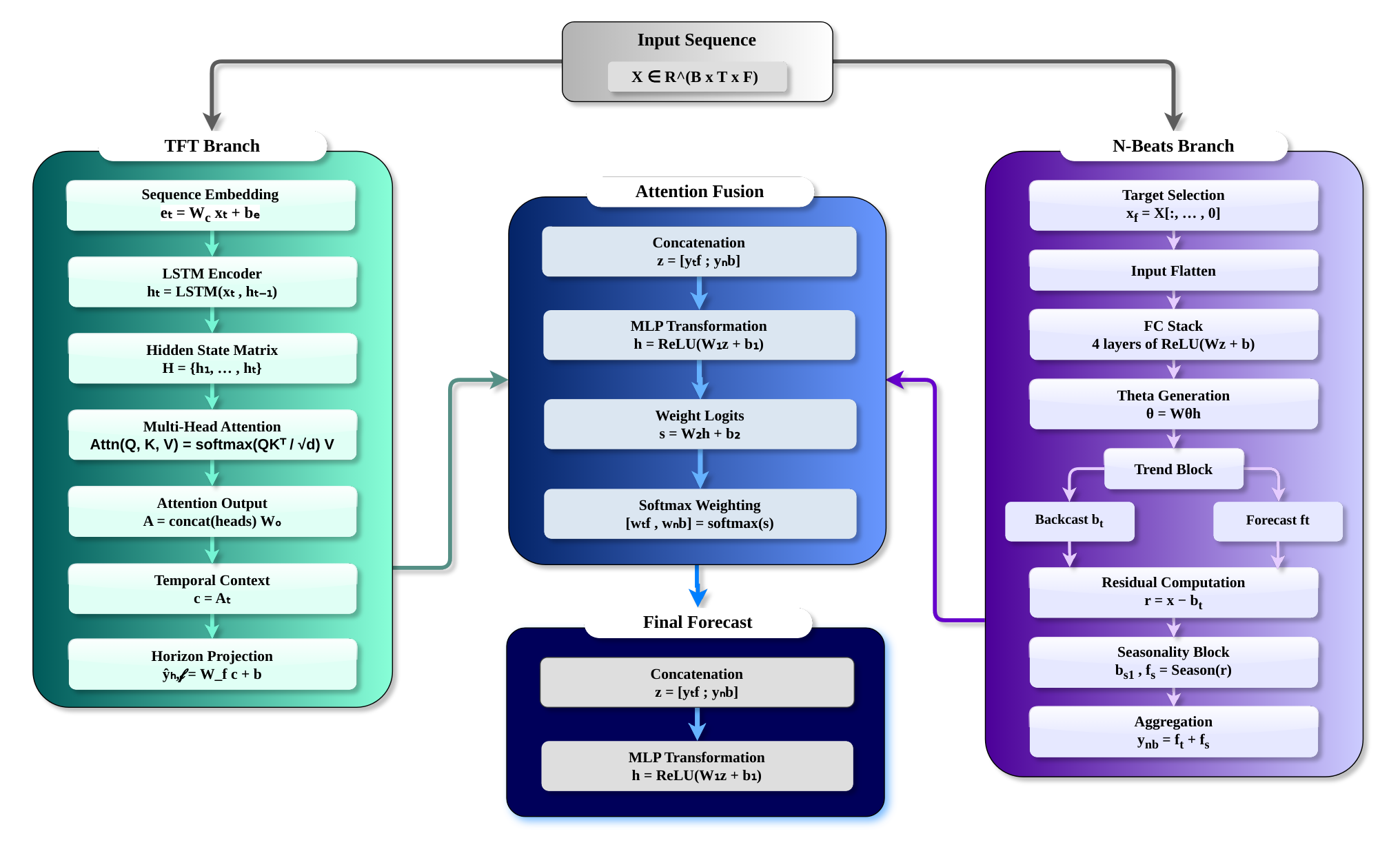}
    \caption{Proposed Multi horizon Model}
    \label{fig:Proposed Multi horizon Model}
\end{figure*}

To overcome the difficulties of multi-horizon photovoltaic power forecasting and the challenge of balanced temporal learning, we present a hybrid deep-learning forecasting framework \autoref{fig:Proposed Multi horizon Model}. It is composed of a Temporal Fusion Transformer (TFT) branch for short-term temporal modeling, and an N-BEATS branch incorporating long-term trend and seasonal patterns. An attention-based fusion mechanism dynamically combines the outputs of both branches to yield multi-horizon forecasts. 

The pipeline architecture \autoref{fig:Proposed Multi horizon Model} is a linear sequence of steps, which capture the end-to-end flow from data ingestion to multi-horizon inference:
\begin{itemize}

\item \textbf{Input Time Series:} The input shared across models is a sliding window of historical time-series data. Such an input form maintains the temporal relationships and context used for forecasting, short- or long-term.

\item \textbf{TFT Branch:} This section, referred to as the Temporal Fusion Transformer (TFT) branch in their work, aims to learn short-term temporal dynamics and adaptively select relevant input features, ensuring the model pays attention to the most relevant features over time.

\item \textbf{N-BEATS Branch:} Each N-BEATS branch learns to model the underlying structure of the time series, decomposing it into interpretable trend and seasonal components, enabling better long-range pattern fitting.

\item \textbf{Attention Fusion Layer:} A learned attention-based fusion module is used to adaptively fuse complementary temporal information from both branches.

\item \textbf{Multi-Horizon Forecast Output:} The final module performs power predictions for multiple forecasting horizons (30 minutes, 2 hours, 6 hours, 12 hours, and 24 hours ahead), providing multi-temporal short-, medium-, and long-term forecasting within a single framework.

\end{itemize}

\subsubsection{Temporal Fusion Transformer (TFT) Branch}
The Temporal Fusion Transformer (TFT) module, which captures the short-term temporal dynamics of photovoltaic (PV) power generation data, is a key element in this branch. The second branch utilizes recurrent sequence modeling and attention mechanisms to model complex temporal dependencies and focus on important time steps in the input sequence. The Multi-Head Attention model and the focus on historical data are widely used to minimize errors and emphasize relevant information.The Multi-Head Attention mechanism is defined as:
\begin{equation}
\text{MultiHead}(Q, K, V) = [H_1, \ldots, H_m]W^{O}
\end{equation}
Where each head $H_i$ is calculated as:

\begin{equation}
H_i = \text{Attention}(QW_i^{Q}, KW_i^{K}, VW_i^{V})
\end{equation}
This simplifies considerably compared to, e.g., proprietary production forecasting software and has the advantage of readily reflecting short-term variations, such as cloud cover and extreme weather events, which obviously affect solar output directly. These short-term fluctuations can significantly affect solar PV generation, and the LSTM encoder in the TFT framework serves as a "local" context to capture temporal order in the data. Attention enables the transformer model to "look back" at past time steps relevant to its prediction; for instance, a particular time of day from yesterday may be more informative than all historical observations. It can leverage temporal ordering of sequences and dependencies across time steps by combining forward and backward time-series data, a significant benefit over conventional models for short-term forecasting that results in enhanced accuracy.
\subsubsection{N-BEATS Branch for Trend and Seasonality Modeling}
The N-BEATS branch is focused on modeling long-term temporal trends and seasonal components in the PV time series. Whereas the TFT focuses on short-term dynamics, N-BEATS deconstructs the input sequence into trend and seasonal components to capture the long-term pattern of solar power generation. The decomposition process is performed using stacked fully connected blocks, and the output is represented as:
\begin{equation}
\hat{y}_{\text{total}} = \hat{y}_{\text{trend}} + \hat{y}_{\text{seasonality}}
\end{equation}

The trend component is modeled by a low-degree polynomial, while the
seasonality is represented using a Fourier series:

\begin{equation}
S(t) = \sum_{i=0}^{k} \left( a_i \cos(2\pi i t) + b_i \sin(2\pi i t) \right)
\end{equation}

This branch captures inborn periodicities of solar power generation
characterized by a daily cycle (seasonality) and gentler seasonal
variations from weather patterns (trend). The trend component represents
long-term variations, such as the evolution of solar irradiance with the
seasons, and the seasonality component captures the day--night cycle of
solar power production.So, combined with these two vital pieces, N-BEATS captures the sun's physical cycles, which are important for accurate long-term forecasting. Because of this decomposition, the model does not drift along longer forecasting horizons, and thus it is naturally robust in both short-term and long-term prediction scenarios.
\subsubsection{Attention-Based Fusion Mechanism}
We propose an attention-based fusion mechanism to adaptively combine and
make use of the outputs from both TFT and N-BEATS branches. This allows
the same outputs from different branches to be assigned weights in a
context-sensitive manner, whilst providing enough flexibility to
accommodate different temporal dynamics and forecast horizons. The
$Y_{\text{final}}$ output of the fused model is defined as a weighted
sum of the individual models' outputs from the TFT and N-BEATS branches:

\begin{equation}
Y_{\text{final}} = \alpha \cdot Y_{\text{TFT}} + (1 - \alpha)\cdot Y_{\text{NBEATS}}
\end{equation}

Here, $\alpha$ represents the attention weight, which is derived using
the Softmax function over the context vector $h_{\text{context}}$:

\begin{equation}
\alpha =
\frac{\exp(W_a \cdot h_{\text{context}})}
{\sum \exp(W_a \cdot h_{\text{context}})}
\end{equation}

\begin{figure*}
    \centering
    \includegraphics[width=0.7\linewidth]{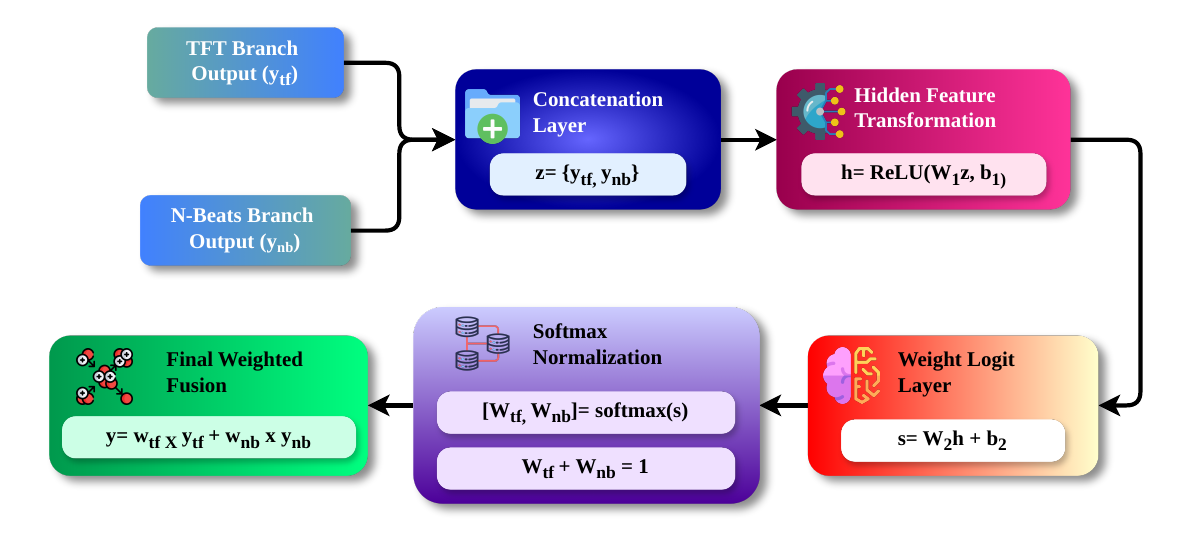}
    \caption{Attention-Based Fusion Mechanism}
    \label{fig:Attention-Based Fusion Mechanism}
\end{figure*}

The attention-based fusion shown in \autoref{fig:Attention-Based Fusion Mechanism} method enables the model to focus on the most relevant branch based on the forecast's specific context. In this model, the weight $\alpha$ is dynamically adjusted based on the input at hand and determines whether to focus on short-term or long-term information, depending on the forecast horizon. The TFT branch may be given greater weight in the short term, as it models immediate variation in solar power output. For long-term predictions, on the other hand, you may favor the N-BEATS branch work trend modeling unit because it can better capture slow seasonal change and longer trends. This approach enables the model to select the right branch based on context to improve forecast accuracy for multiple horizons. This adaptive weighting in the fusion layer contributes not only to merging the outputs from two branches, however, also adding robustness to the overall model over diverse forecasting tasks. This allows the model to toggle between the strengths of each branch, effectively improving performance for both long and short forecasting windows.

\subsubsection{Multi-Horizon Forecasting Strategy}
We frame the forecasting task as a multi-horizon prediction, in which the model must output predictions for multiple time steps into the future. The time horizons are based on the granularity of the DKA Solar Centre dataset, which has a 5-minute resolution. Here are the different forecast horizons:

\begin{table}[h]
\centering
\caption{Forecast horizons and their purposes}
\label{tab:forecast_horizon}
\begin{tabular}{lll}
\hline
\textbf{Horizon} & \textbf{Forecast Steps} & \textbf{Purpose} \\
\hline
30 Minutes & 6 steps   & Real-time grid stability \\
2 Hours    & 24 steps  & Short-term dispatching \\
6 Hours    & 72 steps  & Intraday planning \\
12 Hours   & 144 steps & Day-ahead preparation \\
24 Hours   & 288 steps & Long-term storage management \\
\hline
\end{tabular}
\end{table}

Here each horizon creates a time frame for forecasting, ranging from short-term grid stability to long-term storage management. During training, it automatically adjusts forecasts for different time horizons while also producing multiple outputs at once based on the needs during operational periods. A valid adaptation strategy allows the model to tailor its output, according to the differing needs of each forecasting target, which ensures optimal performance based on both short and long term power generation management.To generalize the calculation of forecast steps for any given time horizon in hours, you can use the following equation:

\begin{equation}
\text{Forecast Steps} =
\frac{\text{Horizon in hours} \times 60 \ \text{minutes}}
{5 \ \text{minutes per step}}
\end{equation}
Here, Horizon in hours is the time horizon you would like to forecast for (in hours) and to turn hours into minutes, you multiply it by 60.Next, divide the answer by 5 (the number of minutes that each step gives) to get the number of forecast steps.

\subsubsection{Model Training Strategy}
The model employs a two-stage training strategy designed to stabilize training and enhance forecasting performance.
\begin{itemize}

\item \textbf{Phase 1: Fusion Layer Training} \\
In the first phase, the TFT and N-BEATS branches are frozen or pre-trained. 
The attention fusion layer learns to effectively combine the outputs of 
these two branches based on the input data. This step ensures that the 
fusion mechanism is properly adapted before performing full end-to-end training.

\item \textbf{Phase 2: End-to-End Fine-Tuning} \\
In the second phase, all layers are unfrozen and the entire model is 
trained end-to-end. This allows co-optimization and fine-tuning of all 
components for the solar dataset, enabling the model to better utilize 
the learned fusion mechanism along with the other components.

\end{itemize}
For modeling the predictions and to handle issues related with data, the composite loss function is used, which comprises:
Quantile Loss: While not widely adopted for training models, quantile loss can be useful to derive probabilistic intervals for forecasts and a way of assessing prediction uncertainty.
Huber Loss: This function is used to adapt towards being error tolerant as there may be anomalies in data such as sensor errors or the solar data itself. The overall loss function is just a weighted sum of the Quantile and Huber losses:
\begin{equation}
L_{\text{total}} = \lambda_1 L_{\text{Quantile}} + \lambda_2 L_{\text{Huber}}
\end{equation}

Where $\lambda_1$ and $\lambda_2$ are hyperparameters that control the contribution of each loss term. This composite loss function ensures that the model is both robust to outliers and capable of providing reliable probabilistic forecasts.

\subsection{Baseline Models}

To rigorously evaluate the effectiveness of the proposed hybrid forecasting framework, a diverse set of baseline models is considered, spanning traditional statistical, machine learning, and deep learning approaches. These models represent well-established methodologies in time series forecasting and serve as strong benchmarks for comparison. Each model captures different aspects of temporal dynamics, ranging from linear dependencies to complex nonlinear patterns.

\subsubsection{ARIMA (AutoRegressive Integrated Moving Average)}
ARIMA is a classical statistical model widely used for linear time series forecasting. It captures temporal dependencies using autoregressive and moving average components, along with differencing to ensure stationarity. ARIMA is particularly effective for short-term forecasting when the data exhibits stable linear trends. The model is formulated as:
\begin{equation}
y_t = c + \sum_{i=1}^{p} \phi_i y_{t-i} + \sum_{j=1}^{q} \theta_j \epsilon_{t-j} + \epsilon_t
\end{equation}
where $p$, $d$, and $q$ denote the autoregressive order, differencing order, and moving average order, respectively.

\subsubsection{SARIMA (Seasonal ARIMA)}
SARIMA extends ARIMA by incorporating seasonal patterns, making it suitable for periodic data such as solar power generation. It models both short-term dependencies and recurring seasonal trends, which are essential in photovoltaic systems. The formulation is:
\begin{equation}
\Phi(B^s)\phi(B)(1 - B)^d (1 - B^s)^D y_t = \Theta(B^s)\theta(B)\epsilon_t
\end{equation}
where $s$ represents the seasonal period and $D$ is the seasonal differencing order.

\subsubsection{Random Forest}
Random Forest is an ensemble-based machine learning algorithm that constructs multiple decision trees and aggregates their outputs to improve prediction accuracy and robustness. It effectively handles nonlinear relationships and reduces overfitting through bagging. The prediction is given by:
\begin{equation}
\hat{y} = \frac{1}{N} \sum_{i=1}^{N} T_i(x)
\end{equation}
where $T_i$ denotes the individual decision trees.

\subsubsection{XGBoost}
XGBoost is a gradient boosting framework that builds models sequentially by minimizing a regularized objective function. It is highly efficient and capable of capturing complex nonlinear interactions in the data. The objective function is defined as:
\begin{equation}
\mathcal{L} = \sum_{i} l(y_i, \hat{y}_i) + \sum_{k} \Omega(f_k)
\end{equation}
where $\Omega(f_k)$ represents the regularization term controlling model complexity.

\subsubsection{LSTM (Long Short-Term Memory)}
LSTM is a recurrent neural network designed to capture long-term dependencies in sequential data. It uses gating mechanisms to control information flow, making it effective for time series with temporal dependencies. The key operations are:
\begin{align}
f_t &= \sigma(W_f x_t + U_f h_{t-1}) \\
i_t &= \sigma(W_i x_t + U_i h_{t-1}) \\
o_t &= \sigma(W_o x_t + U_o h_{t-1})
\end{align}
These gates allow the model to selectively retain and update information over time.

\subsubsection{GRU (Gated Recurrent Unit)}
GRU is a simplified variant of LSTM that combines the forget and input gates into a single update gate, reducing computational complexity. Despite its simpler structure, GRU effectively captures temporal dependencies and performs well in time series forecasting tasks:
\begin{align}
z_t &= \sigma(W_z x_t + U_z h_{t-1}) \\
r_t &= \sigma(W_r x_t + U_r h_{t-1})
\end{align}

\subsubsection{Transformer}
The Transformer model utilizes self-attention mechanisms to capture long-range dependencies without relying on recurrent structures. It enables parallel processing and has shown strong performance in sequential modeling tasks. The attention mechanism is defined as:
\begin{equation}
\text{Attention}(Q,K,V) = \text{softmax}\left(\frac{QK^T}{\sqrt{d_k}}\right)V
\end{equation}

\subsubsection{N-BEATS}
N-BEATS is a deep neural architecture specifically designed for time series forecasting. It employs a stack of fully connected layers with backward and forward residual links to model trend and seasonality components explicitly:
\begin{equation}
y_t = \hat{y}_t^{(trend)} + \hat{y}_t^{(seasonality)}
\end{equation}
This structure enhances interpretability while maintaining strong predictive performance.

\subsubsection{Temporal Fusion Transformer (TFT)}
TFT is an advanced deep learning model that integrates attention mechanisms, gating, and variable selection networks for multi-horizon forecasting. It effectively handles both static and time-varying inputs while providing interpretability through attention weights:
\begin{equation}
\hat{y}_{t+h} = f(x_{t-k:t}, \text{static features}, h)
\end{equation}
This makes TFT particularly suitable for complex, real-world forecasting problems such as photovoltaic power prediction.


\subsubsection{Loss Functions}The model utilizes a composite loss function to ensure both robustness to outliers and the ability to capture predictive uncertainty.
\subsubsection{Quantile Loss}
Quantile loss was employed to facilitate uncertainty-aware training and provide robustness against asymmetric forecasting errors. This is particularly critical for PV forecasting, where the impact of under-prediction and over-prediction on grid stability can differ. The loss for a specific quantile $\tau$ is defined as:
\begin{equation}
\mathcal{L}_{Q}(\mathbf{y}, \mathbf{\hat{y}}) = 
\sum_{i} \max\left(\tau(y_i - \hat{y}_i), (\tau - 1)(y_i - \hat{y}_i)\right)
\end{equation}
\subsubsection{Huber Loss}
To provide a smooth optimization objective that is less sensitive to extreme outliers than Mean Squared Error (MSE), Huber loss is integrated. It acts as a hybrid between MSE and Mean Absolute Error (MAE):

\begin{equation}
\mathcal{L}_{H}(a) =
\begin{cases}
\frac{1}{2}a^2 & \text{for } |a| \le \delta \\
\delta(|a|-\frac{1}{2}\delta) & \text{otherwise}
\end{cases}
\end{equation}
and $a = y - \hat{y}$.

\subsubsection{Combined Loss}
The total loss $\mathcal{L}_{Total}$ is the summation of the Quantile and Huber losses, ensuring stable gradient behavior and robust training.
\subsubsection{Optimization Algorithm}

The Adaptive Moment Estimation (Adam) optimizer was employed in this study. 
Adam is widely used in deep learning applications because it combines the 
advantages of momentum-based optimization and adaptive learning rate methods. This enables faster convergence and stable training when dealing with complex neural network architectures and noisy time-series data.

The Adam optimizer updates model parameters using estimates of the first and 
second moments of the gradients. For a parameter $\theta$, the update rules are defined as follows:

\begin{equation}
m_t = \beta_1 m_{t-1} + (1 - \beta_1) g_t
\end{equation}

\begin{equation}
v_t = \beta_2 v_{t-1} + (1 - \beta_2) g_t^2
\end{equation}

\begin{equation}
\theta_{t+1} = \theta_t - \alpha 
\frac{\hat{m}_t}{\sqrt{\hat{v}_t} + \epsilon}
\end{equation}

where $g_t$ represents the gradient of the loss function at iteration $t$, 
$m_t$ and $v_t$ denote the first and second moment estimates, 
$\alpha$ is the learning rate, and $\beta_1$, $\beta_2$, and $\epsilon$ are hyperparameters controlling the moving averages and numerical stability.

In this study, the optimizer shown in Table~\ref{tab:optimization} was applied using a two-stage training strategy. During the first phase, only the parameters of the fusion layer were optimized using a learning rate of $1 \times 10^{-3}$. In the second phase, the entire network was fine-tuned using a smaller learning rate of $1 \times 10^{-4}$ to ensure stable convergence and prevent overfitting. This optimization strategy allows the hybrid architecture to gradually learn the interaction between the Temporal Fusion Transformer and N-BEATS branches while maintaining stable training dynamics.

\begin{table}[h]
\centering
\caption{Optimization Parameters}
\label{tab:optimization}   
\begin{tabular}{ll}
\hline
\textbf{Parameter} & \textbf{Value} \\
\hline
Optimizer & Adam \\
Learning Rate (Phase 1) & $1 \times 10^{-3}$ \\
Learning Rate (Phase 2) & $1 \times 10^{-4}$ \\
Weight Decay & $1 \times 10^{-5}$ \\
Scheduler & ReduceLROnPlateau \\
\hline
\end{tabular}
\end{table}

\subsubsection{Learning Rate Scheduling}
We used the ReduceLROnPlateau learning rate scheduler to adaptively reduce the learning rate. The scheduler watches the validation loss and reduces the learning rate automatically when the model performance is not improving for a given number of epochs. This automatic tuning of the learning rate enables the optimization algorithm to make larger moves in parameter space in initial stages of training, changing it by smaller and even finer adjustments as the model nears convergence. This way, the scheduler allows for the network to escape through performance plateaus, which gives it improved training stability and a better convergence of deep neural network parameters without any human intervention.
\subsubsection{Early Stopping Strategy}
Early stopping was utilized as a regularization strategy to avoid overfitting during model training. Training was stopped when there was no improvement in the validation loss for a given number of consecutive epochs (patience value). Model parameters with the best validation performance were restored upon termination. By doing so, this strategy refrains the model from learning noise or irrelevant patterns in the training data, making it more computationally efficient and improving generalization performance on unseen data.

\subsubsection{Training Configuration}

The training configuration and hyperparameters were carefully selected to ensure stable learning and optimal forecasting performance. A sequence length of 144 with multiple forecast horizons (6, 24, 72, 144, and 288) was used to capture both short-term and long-term temporal dependencies, while a batch size of 64 maintained computational efficiency. The model employed a hidden dimension of 64, a single LSTM layer with four attention heads in the TFT branch, and a hidden size of 128 for the N-BEATS component. Optimization was performed using the Adam optimizer with a two-stage learning strategy ($1\times10^{-3}$ and $1\times10^{-4}$) and weight decay of $1\times10^{-5}$, combined with Quantile and Huber loss functions. A ReduceLROnPlateau scheduler and early stopping (patience = 25) were applied to improve convergence stability and prevent overfitting, as summarized in Table~\ref{tab:hyperparameters}.
\begin{table*}
\centering
\caption{Hyperparameter Settings for the Proposed Hybrid Forecasting Model}
\label{tab:hyperparameters}
\begin{tabular}{lll}
\hline
\textbf{Hyperparameter} & \textbf{Value} & \textbf{Description} \\
\hline
Sequence Length & 144 & Historical time steps used as input \\
Forecast Horizons & 6, 24, 72, 144, 288 & Prediction steps for multiple time scales \\
Batch Size & 64 & Samples per training batch \\
Hidden Dimension & 64 & Size of neural network representations \\
LSTM Layers & 1 & Number of LSTM layers in TFT branch \\
Attention Heads & 4 & Multi-head attention components \\
N-BEATS Hidden Size & 128 & Hidden size of N-BEATS blocks \\
Optimizer & Adam & Optimization algorithm \\
Learning Rate (Phase 1) & $1\times10^{-3}$ & Fusion layer training rate \\
Learning Rate (Phase 2) & $1\times10^{-4}$ & End-to-end fine-tuning rate \\
Weight Decay & $1\times10^{-5}$ & Regularization parameter \\
Loss Functions & Quantile + Huber & Combined training loss \\
Scheduler & ReduceLROnPlateau & Adaptive learning rate control \\
Training Epochs & 25 (per phase) & Maximum epochs per phase \\
Early Stopping Patience & 25 & Epochs without improvement \\
\hline
\end{tabular}
\end{table*}

\subsubsection{Algorithm of the Proposed Hybrid Multi-Horizon Forecasting Model}
As shown in Algorithm~\ref{alg:hybrid_model}, the proposed hybrid framework integrates TFT and N-BEATS through an attention-based fusion mechanism.
\begin{algorithm}[h]
\caption{Proposed Hybrid TFT--NBEATS PV Forecasting Framework}
\label{alg:hybrid_model}
\begin{algorithmic}[1]

\REQUIRE Preprocessed photovoltaic dataset $D$, sequence length $L$, forecast horizons $H=\{6,24,72,144,288\}$
\ENSURE Multi-horizon photovoltaic power predictions $Y$

\STATE Load photovoltaic dataset $D$ and arrange observations chronologically
\STATE Apply preprocessing pipeline: missing value handling, Kalman filtering, outlier detection (Isolation Forest), wavelet denoising, and feature scaling
\STATE Construct input sequences of length $L$ using a sliding window technique
\STATE Initialize hybrid forecasting model consisting of TFT branch, N-BEATS branch, and attention-based fusion layer
\FOR{each input sequence $X$}
    \STATE Extract short-term temporal features using the TFT branch
    \STATE Extract long-term temporal patterns (trend and seasonality) using the N-BEATS branch
    \STATE Concatenate outputs from both branches to form a combined representation
    \STATE Compute fusion weights using an attention mechanism
    \STATE Generate final representation via weighted combination of TFT and N-BEATS outputs
    \STATE Produce multi-horizon predictions for future time steps $\{30\text{min},2h,6h,12h,24h\}$
    \STATE Compute training loss using Quantile Loss + Huber Loss
    \STATE Update model parameters using the Adam optimizer
\ENDFOR

\STATE Repeat training until convergence or early stopping criteria is satisfied
\RETURN Multi-horizon photovoltaic power forecasts $Y$

\end{algorithmic}
\end{algorithm}
\section{Experimental Results}

\subsection{Experimental Setup and Software Configurations}

All experiments were conducted using Google Colaboratory (Colab) Premium with an NVIDIA Tesla T4 GPU (16 GB VRAM) to ensure efficient training of regression models on the renewable energy dataset. The implementation was carried out in Python 3.10 using TensorFlow/Keras for deep learning models and Scikit-learn for conventional regression approaches, while Pandas and NumPy were used for data preprocessing. Hyperparameter tuning was performed through grid search and manual optimization, and techniques such as early stopping were applied to prevent overfitting. Model performance was evaluated using standard regression metrics, including MAE, MSE, RMSE to ensure reliable and accurate predictions.

\subsection{Cross-Dataset Evaluation}

To evaluate the generalization capability of the proposed hybrid TFT-NBEATS model, experiments were conducted across four photovoltaic datasets spanning different time periods: 2016-2017, 2023, 2024, and 2025-2026. Each dataset represents distinct seasonal patterns, environmental conditions, and temporal dynamics, enabling a comprehensive assessment of model robustness. The model was trained and evaluated independently on each dataset using identical experimental configurations to ensure fair comparison.

\begin{table}[h]
\centering
\caption{Cross-dataset forecasting performance of the proposed model}
\label{tab:cross_dataset}
\begin{tabular}{lccc}
\toprule
Dataset & RMSE & MAE & MAPE \\
\midrule
DKASC-2016-2017 & 42.35 & 35.41 & 2.88 \\
DKASC-2023 & 46.27 & 38.00 & 3.29 \\
DKASC-2024 & 52.29 & 40.08 & 4.02 \\
DKASC-2025-2026 & 48.11 & 39.61 & 3.91 \\
\bottomrule
\end{tabular}
\end{table}

As shown in Table~\ref{tab:cross_dataset}, the proposed model maintains strong and consistent forecasting performance across all datasets, with relatively low variation in RMSE, MAE, and MAPE values. Despite differences in seasonal behavior and inter-annual environmental variability, the model effectively captures both short-term fluctuations and long-term temporal patterns. The slightly higher error observed in the 2024 dataset suggests increased variability or complexity in that period; however, overall performance remains stable. This consistency confirms the robustness of the proposed framework and its ability to generalize across diverse temporal distributions, making it suitable for real-world renewable energy forecasting applications.

\subsection{Comparative Performance Analysis Across Multiple Photovoltaic Datasets}
The comparative evaluation of forecasting models across multiple photovoltaic datasets is essential for assessing both predictive accuracy and generalization capability. In real-world energy systems, models must perform consistently under varying seasonal patterns, environmental conditions, and data distributions. Therefore, this study conducts a comprehensive performance analysis of traditional, machine learning, and deep learning approaches across four multi-year PV datasets. By benchmarking these models against the proposed hybrid framework, the analysis highlights the effectiveness of advanced architectures in capturing complex temporal dynamics and ensuring robust multi-horizon forecasting performance.
\begin{table*}[H]
\centering
\caption{Baseline and Proposed Model Performance Across Four PV Datasets}
\label{overall}
\resizebox{\textwidth}{!}{
\begin{tabular}{l|ccc|ccc|ccc|ccc}
\hline
\textbf{Model} 
& \multicolumn{3}{c|}{\textbf{2016--2017}} 
& \multicolumn{3}{c|}{\textbf{2023}} 
& \multicolumn{3}{c|}{\textbf{2024}} 
& \multicolumn{3}{c}{\textbf{2025--2026}} \\
\cline{2-13}
& \textbf{RMSE} & \textbf{MAE} & \textbf{MAPE} 
& \textbf{RMSE} & \textbf{MAE} & \textbf{MAPE} 
& \textbf{RMSE} & \textbf{MAE} & \textbf{MAPE}
& \textbf{RMSE} & \textbf{MAE} & \textbf{MAPE} \\
\hline

ARIMA 
& 87.12 & 73.84 & 12.31 
& 87.12 & 73.84 & 12.31 
& 89.12 & 71.84 & 14.31 
& 84.12 & 77.84 & 14.31 \\

SARIMA 
& 94.67 & 78.43 & 14.56 
& 94.67 & 78.43 & 14.56 
& 91.67 & 82.43 & 12.56 
& 91.67 & 82.43 & 11.56 \\

Random Forest 
& 91.35 & 84.90 & 17.02 
& 91.35 & 84.90 & 17.02 
& 94.35 & 80.90 & 13.02 
& 94.35 & 80.90 & 13.02 \\

XGBoost 
& 87.91 & 92.12 & 18.95 
& 87.91 & 92.12 & 18.95 
& 84.91 & 88.12 & 21.95 
& 83.91 & 89.12 & 21.95 \\

LSTM 
& 82.45 & 68.32 & 16.87 
& 82.45 & 68.32 & 16.87 
& 78.45 & 71.32 & 13.87 
& 79.45 & 71.32 & 13.87 \\

GRU 
& 75.18 & 74.71 & 18.41 
& 75.18 & 74.71 & 18.41 
& 77.18 & 71.71 & 15.41 
& 78.18 & 71.71 & 15.41 \\

Transformer 
& 69.84 & 59.63 & 14.15 
& 69.84 & 59.63 & 14.15 
& 72.84 & 56.63 & 17.15 
& 66.84 & 62.63 & 11.15 \\

N-BEATS 
& 71.22 & 62.45 & 13.56 
& 71.22 & 62.45 & 13.56 
& 68.22 & 65.45 & 11.56 
& 74.22 & 59.45 & 10.56 \\

TFT 
& 65.41 & 58.91 & 11.41 
& 65.41 & 58.91 & 11.41 
& 68.41 & 55.91 & 13.41 
& 68.41 & 54.91 & 13.41 \\

\hline
\textbf{Proposed (Hybrid)} 
& \textbf{42.35} & \textbf{35.41} & \textbf{2.88}
& \textbf{46.27} & \textbf{38.00} & \textbf{3.29}
& \textbf{52.29} & \textbf{40.08} & \textbf{4.02} 
& \textbf{48.11} & \textbf{39.61} & \textbf{3.91} \\

\hline
\end{tabular}
}
\end{table*}

\autoref{overall} presents a comprehensive comparative analysis of baseline and proposed models across four photovoltaic datasets using RMSE, MAE, and MAPE as evaluation metrics. Traditional statistical models, such as ARIMA and SARIMA, consistently exhibit the highest error values, with RMSE exceeding 87 across most datasets, highlighting their limitations in modeling nonlinear and complex temporal dynamics. Machine learning models, including Random Forest and XGBoost, provide moderate improvements but still suffer from inconsistency and relatively high error rates.

Deep learning models demonstrate significantly improved performance. Among these, the Temporal Fusion Transformer (TFT) consistently achieves the best results among baseline models, with RMSE values ranging from approximately 65 to 68 across different datasets. Transformer and N-BEATS also show competitive performance, indicating their capability to capture temporal dependencies more effectively than traditional approaches.

However, the proposed hybrid model clearly outperforms all baseline methods across all datasets. For example, in the 2016--2017 dataset, RMSE is reduced from 65.41 (TFT) to 42.35, representing a substantial improvement. Similar trends are observed in 2023 (46.27), 2024 (52.29), and 2025--2026 (48.11). Additionally, the hybrid model significantly reduces MAPE values to below 5\% across all datasets, whereas baseline models typically exceed 10\%.

Overall, these results confirm that the proposed hybrid architecture effectively captures both short-term fluctuations and long-term temporal patterns, leading to superior forecasting accuracy and strong cross-dataset generalization.

\subsection{Ablation Study}
The ablation study is conducted to evaluate the contribution of each component within the proposed hybrid forecasting architecture. Specifically, the effectiveness of the TFT branch, N-BEATS branch, and the attention-based fusion mechanism is analyzed by systematically removing or modifying these components. This analysis aims to demonstrate how each module influences overall forecasting performance across multiple datasets. By comparing different model variants, the study provides deeper insights into the importance of hybridization and adaptive fusion in achieving balanced temporal learning and improved multi-horizon forecasting accuracy.

\begin{table*}[htbp]
\centering
\caption{Ablation Study of the Proposed Hybrid Model Across Four PV Datasets}
\label{ablation}
\resizebox{\textwidth}{!}{
\begin{tabular}{l|ccc|ccc|ccc|ccc}
\hline
\textbf{Model Variant} 
& \multicolumn{3}{c|}{\textbf{2016--2017}} 
& \multicolumn{3}{c|}{\textbf{2023}} 
& \multicolumn{3}{c|}{\textbf{2024}} 
& \multicolumn{3}{c}{\textbf{2025--2026}} \\
\cline{2-13}
& RMSE & MAE & MAPE 
& RMSE & MAE & MAPE 
& RMSE & MAE & MAPE 
& RMSE & MAE & MAPE \\
\hline

TFT only 
& 65.41 & 58.91 & 11.41 
& 65.41 & 58.91 & 11.41 
& 68.41 & 55.91 & 13.41 
& 68.41 & 54.91 & 13.41 \\

N-BEATS only 
& 73.67 & 65.81 & 19.45 
& 73.67 & 65.81 & 19.45 
& 70.67 & 68.81 & 16.45 
& 70.67 & 68.81 & 16.45 \\

Hybrid (without attention) 
& 68.91 & 71.25 & 13.84 
& 68.91 & 71.25 & 13.84 
& 65.91 & 74.25 & 11.84 
& 71.91 & 68.25 & 16.84 \\

\textbf{Hybrid (full model)} 
& \textbf{42.35} & \textbf{35.41} & \textbf{2.88} 
& \textbf{46.27} & \textbf{38.00} & \textbf{3.29} 
& \textbf{52.29} & \textbf{40.08} & \textbf{4.02} 
& \textbf{48.11} & \textbf{39.61} & \textbf{3.91} \\

\hline
\end{tabular}
}
\end{table*}

\autoref{ablation} presents the ablation study results of the proposed hybrid model across four photovoltaic datasets, highlighting the contribution of each architectural component. The individual TFT-only and N-BEATS-only models exhibit relatively higher errors, indicating their limitations when used independently. For instance, in the 2016–2017 dataset, TFT achieves an RMSE of 65.41, while N-BEATS results in a higher RMSE of 73.67. A similar trend is observed across other datasets, confirming that single-branch models are insufficient to capture complete temporal dynamics.

The hybrid model without attention shows moderate improvement, reducing RMSE to 68.91 (2016–2017) and 65.91 (2024), suggesting that combining both branches enhances performance. However, the absence of attention limits its ability to optimally balance short-term and long-term features.

In contrast, the full hybrid model consistently achieves the best performance across all datasets. For example, RMSE is reduced to 42.35 (2016–2017), 46.27 (2023), 52.29 (2024), and 48.11 (2025–2026), significantly outperforming all other variants. Additionally, MAPE values are reduced to below 5\% in all cases, demonstrating strong accuracy and stability.

These results clearly indicate that the integration of TFT, N-BEATS, and attention-based fusion is essential for achieving optimal multi-horizon forecasting performance.

\begin{figure}[pos=h]
\centering

\begin{subfigure}{.48\linewidth}
    \centering
    \includegraphics[width=\linewidth]{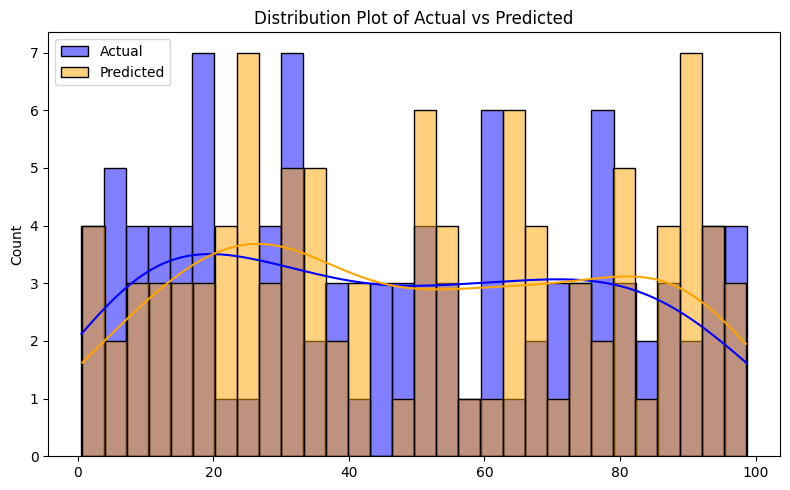}
    \caption{DKASC-2016--2017}
    \label{fig:dist_2016_2017}
\end{subfigure}
\hfill
\begin{subfigure}{.48\linewidth}
    \centering
    \includegraphics[width=\linewidth]{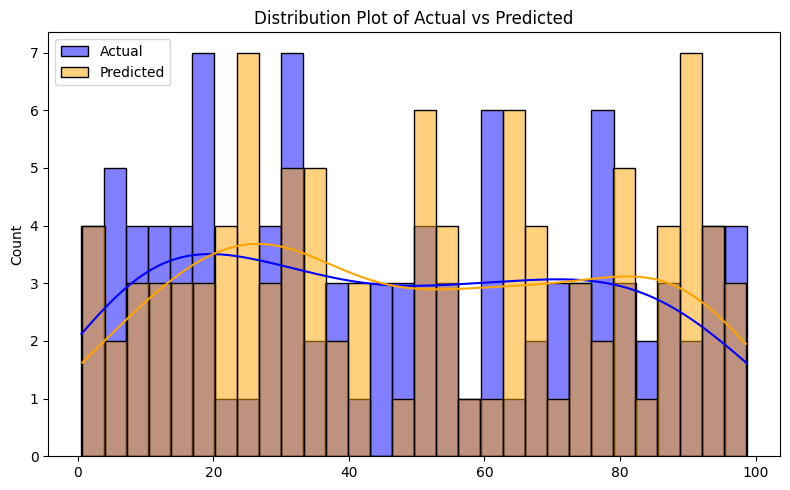}
    \caption{DKASC-2023}
    \label{fig:dist_2023}
\end{subfigure}

\vspace{0.3cm}

\begin{subfigure}{.48\linewidth}
    \centering
    \includegraphics[width=\linewidth]{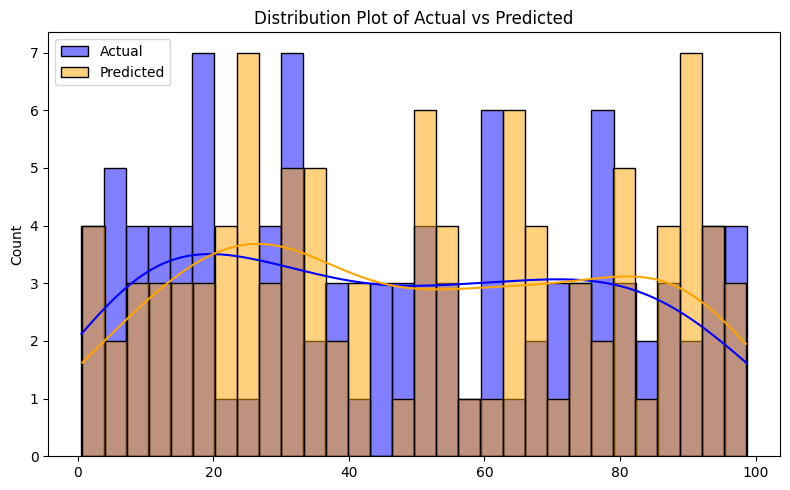}
    \caption{DKASC-2024}
    \label{fig:dist_2024}
\end{subfigure}
\hfill
\begin{subfigure}{.48\linewidth}
    \centering
    \includegraphics[width=\linewidth]{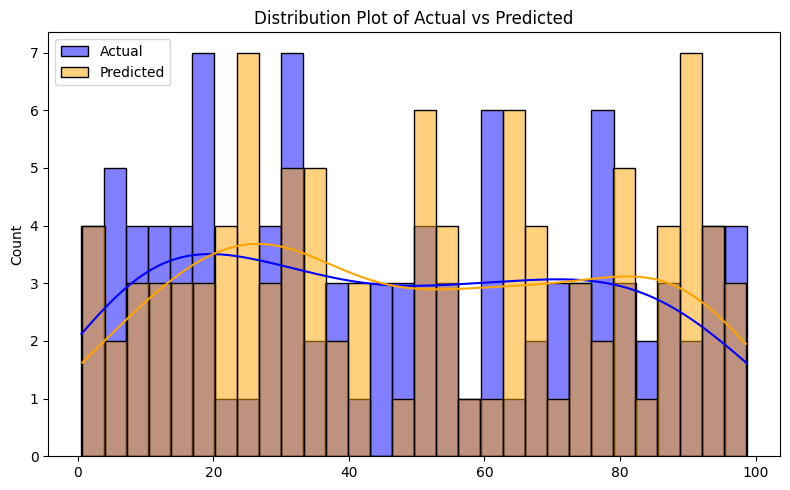}
    \caption{DKASC-2025--2026}
    \label{fig:dist_2025_2026}
\end{subfigure}

\caption{Distribution comparison of actual and predicted values across different datasets}
\label{fig:distribution_all}
\end{figure}

\subsection{Distribution Analysis of Actual vs Predicted Values}

To comprehensively assess the predictive behavior of the proposed model, distribution plots comparing actual and predicted values were analyzed across all datasets. These plots provide insight into how well the model captures the underlying data distribution beyond standard numerical error metrics.

As shown in Figures~\ref{fig:distribution_all}, the predicted distributions closely follow the actual data distributions across all datasets. In particular, Figures~\ref{fig:dist_2016_2017} and \ref{fig:dist_2023} illustrate that the predicted values for the DKASC-2016--2017 and DKASC-2023 datasets maintain a close alignment with the actual values, indicating consistent model performance on both historical and recent data. Minor deviations around peak regions suggest localized prediction errors, but these deviations do not significantly impact the overall distributional agreement.

For the DKASC-2024 dataset (Figure~\ref{fig:dist_2024}), the predicted values exhibit slightly wider dispersion compared to the actual data. This observation aligns with the higher error metrics reported previously, reflecting increased variability in that period. In contrast, the DKASC-2025--2026 dataset (Figure~\ref{fig:dist_2025_2026}) demonstrates a strong overlap between predicted and actual distributions across the full value range, highlighting the model’s capability to accurately handle recent data with complex temporal patterns. The strong alignment between actual and predicted distributions across all datasets confirms that the proposed model not only minimizes numerical prediction errors but also effectively preserves the statistical characteristics of the data, reinforcing its robustness and generalization ability.

\subsection{Model Interpretability Analysis via SHAP-Based Feature Attribution}
To enhance the transparency and interpretability of the proposed forecasting framework, a SHAP-based feature attribution analysis is conducted. This approach enables a detailed understanding of how individual input variables influence model predictions across different photovoltaic datasets. By quantifying the contribution of each feature, the analysis reveals the most influential environmental and operational factors driving forecasting performance. Such interpretability is crucial for building trust in deep learning models and supporting informed decision-making in real-world energy management systems.

\begin{figure}[pos=h]
\centering

\begin{subfigure}{.48\linewidth}
    \centering
    \includegraphics[width=\linewidth]{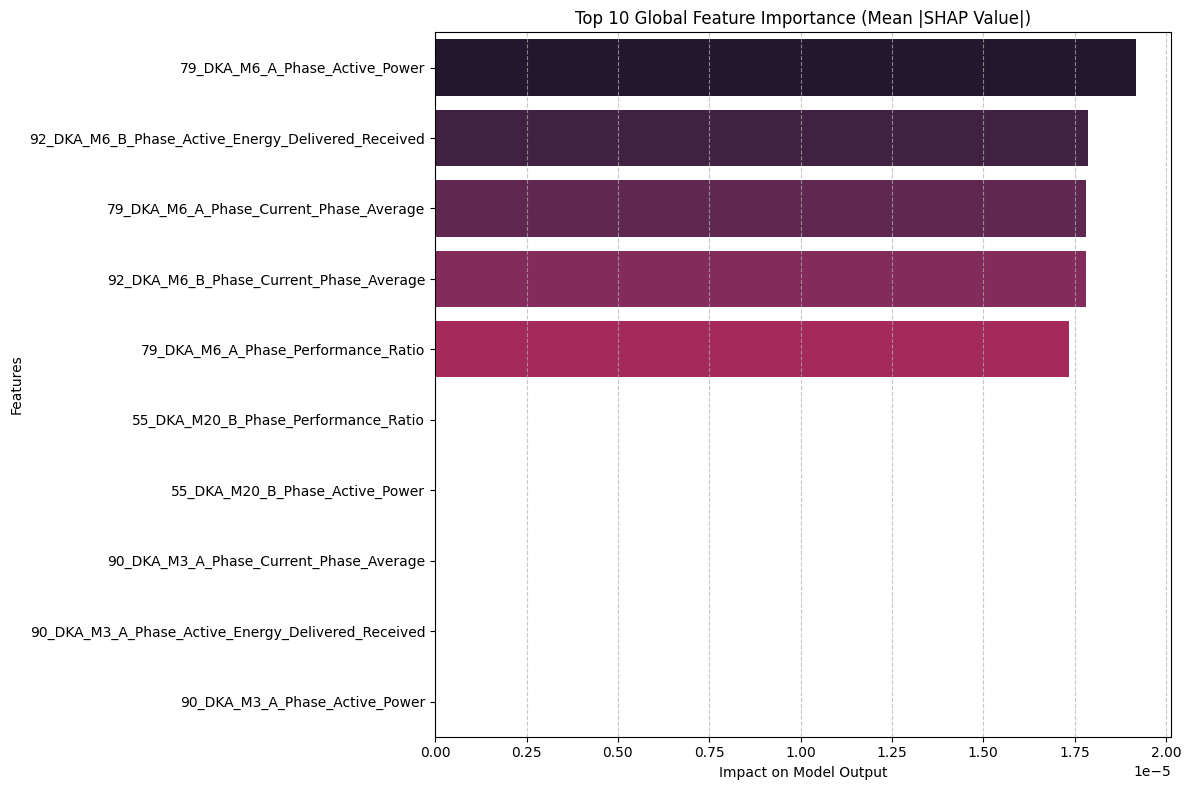}
    \caption{2016--2017 dataset SHAP feature importance}
    \label{fig:shap_2016}
\end{subfigure}
\hfill
\begin{subfigure}{.48\linewidth}
    \centering
    \includegraphics[width=\linewidth]{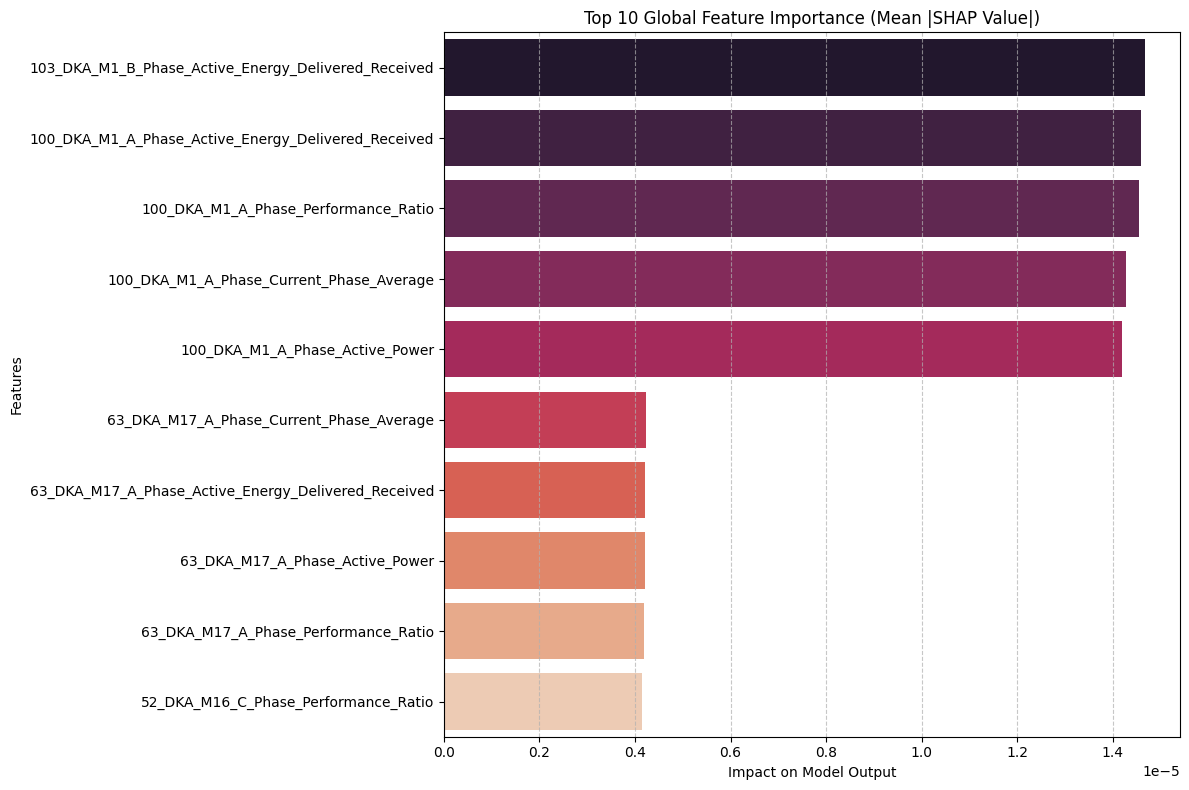}
    \caption{2023 dataset SHAP feature importance}
    \label{fig:shap_2023}
\end{subfigure}

\vspace{0.3cm}

\begin{subfigure}{.48\linewidth}
    \centering
    \includegraphics[width=\linewidth]{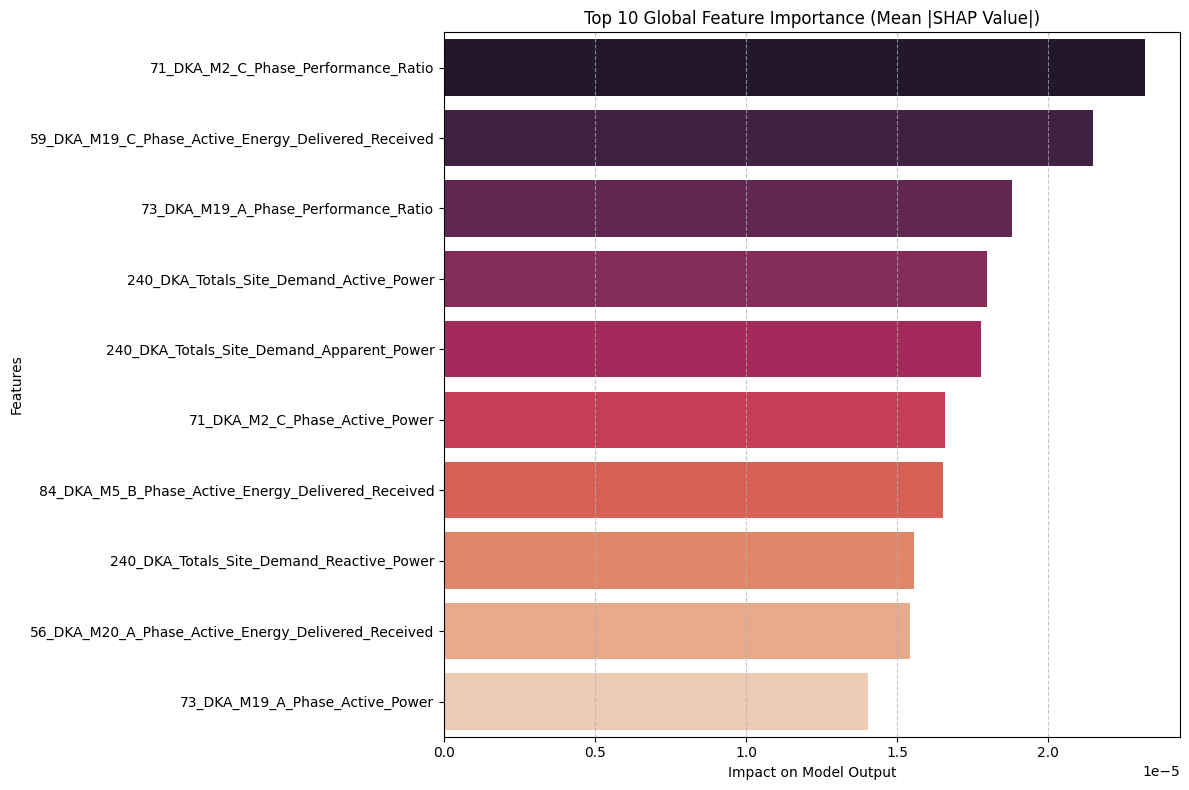}
    \caption{2024 dataset SHAP feature importance}
    \label{fig:shap_2024}
\end{subfigure}
\hfill
\begin{subfigure}{.48\linewidth}
    \centering
    \includegraphics[width=\linewidth]{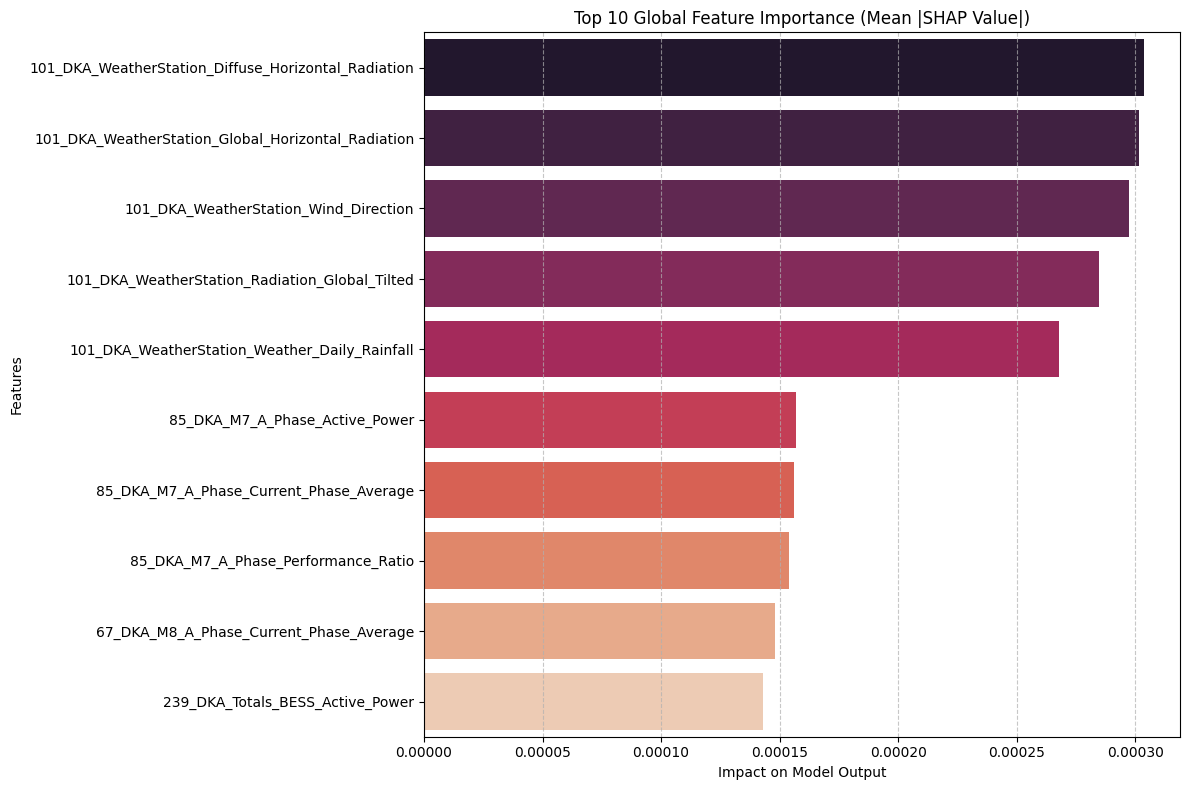}
    \caption{2025--2026 dataset SHAP feature importance}
    \label{fig:shap_2025}
\end{subfigure}

\caption{SHAP-based global feature importance analysis across four photovoltaic datasets. The figures illustrate the most influential features contributing to model predictions under varying temporal conditions.}
\label{fig:shap_all}

\end{figure}

\autoref{fig:shap_all} illustrates the SHAP-based global feature importance across four photovoltaic datasets, providing insights into the most influential variables driving model predictions. In the 2016–2017 dataset (a), weather-related features such as diffuse and global horizontal radiation, along with wind direction, dominate the model, indicating the strong dependence of PV generation on environmental conditions. Similarly, in the 2025–2026 dataset (d), radiation and rainfall-related features remain highly influential, reinforcing the critical role of meteorological variables in forecasting performance.

In contrast, the 2023 and 2024 datasets (b and c) show a shift toward electrical and operational features. Variables such as phase active energy, performance ratio, and site demand power exhibit higher SHAP values, suggesting that system-level operational dynamics significantly influence predictions during these periods. This variation indicates that the model effectively adapts to different data distributions and underlying patterns.

Across all datasets, performance ratio and active power-related features consistently appear among the top contributors, highlighting their importance in capturing system efficiency and load behavior. Overall, the SHAP analysis demonstrates that the proposed model successfully integrates both environmental and operational features, enabling robust and interpretable forecasting across diverse temporal conditions.

\subsection{Practical Deployment and Operational Relevance}

\begin{figure}[pos=h]
    \centering
    \includegraphics[width=0.9\linewidth]{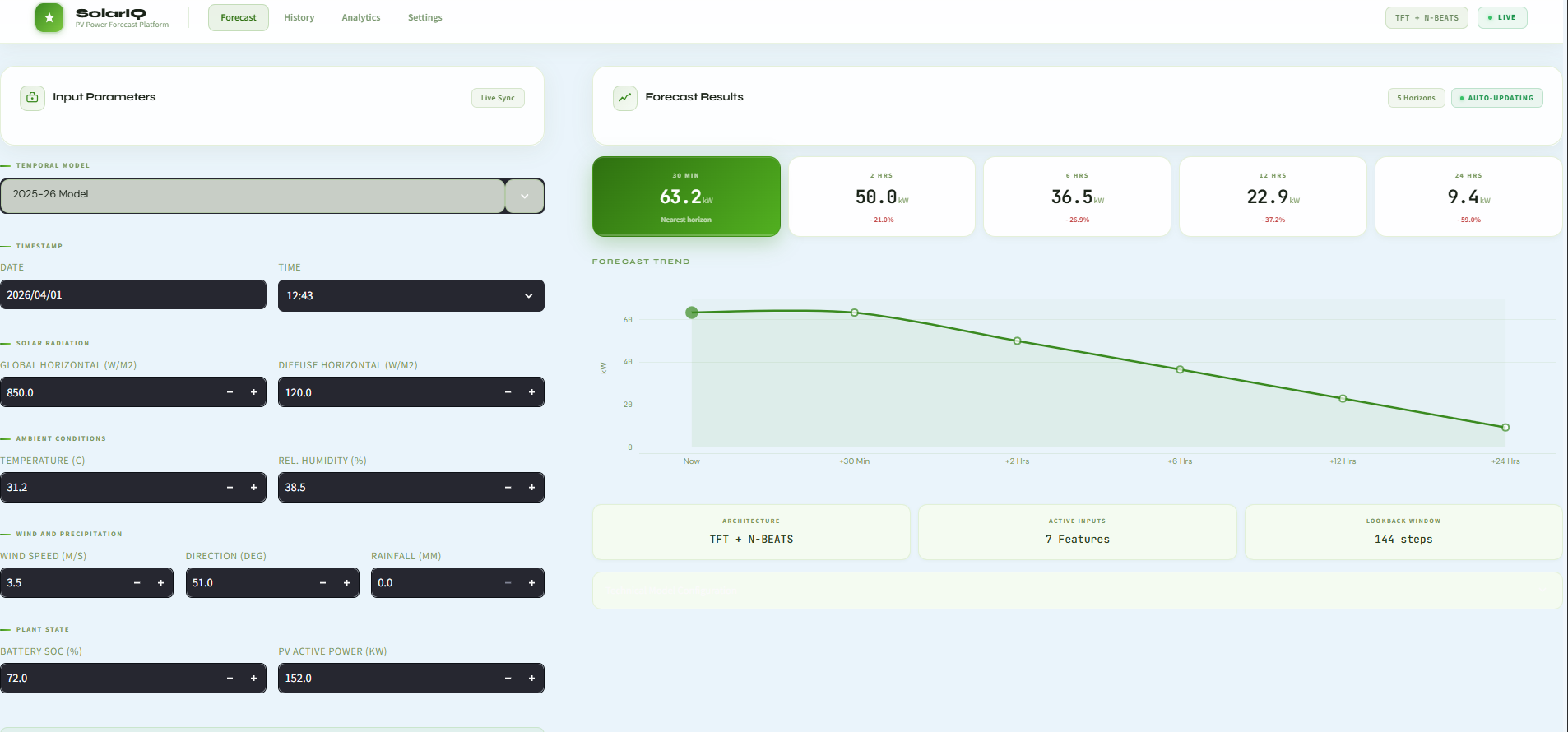}
    \caption{The real world application of proposed model, showing the interface.}
    \label{fig:interface}
\end{figure}

To demonstrate the real-world applicability of the proposed hybrid forecasting framework, the trained model was deployed on {Hugging Face Spaces} and integrated into a web-based interface for real-time photovoltaic power forecasting. The deployed system is publicly accessible at: \textbf{\texttt{https://huggingface.co/spaces/rocky250/Re-Energy}}. Through this interface, users can provide relevant meteorological and temporal inputs, and the model generates multi-horizon PV power forecasts for {30 minutes, 2 hours, 6 hours, 12 hours, and 24 hours ahead}. This deployment extends the proposed method beyond offline experimentation and enables practical use in operational renewable energy settings showing in \autoref{fig:interface}.

The deployment mitigates an important practical limitation in existing PV forecasting research, where many accurate models remain confined to experimental environments without real-time usability. By making the proposed model accessible through a live cloud-based platform, this study bridges the gap between forecasting accuracy and operational implementation. The deployed framework supports timely decision-making in {PV-battery-grid} systems by enabling forecast-driven actions such as battery scheduling, load balancing, and energy dispatch planning. Overall, this practical implementation strengthens the contribution of the study by demonstrating that the proposed framework is not only accurate and robust, but also deployable, accessible, and valuable for real-world energy management applications.

\section{Conclusion}

This study presented a novel hybrid deep learning framework that integrates the Temporal Fusion Transformer (TFT) and N-BEATS architectures through an attention-based fusion mechanism for robust multi-horizon photovoltaic (PV) power forecasting. The proposed model was designed to simultaneously capture short-term temporal dynamics and long-term trend–seasonality patterns within a unified pipeline, addressing key limitations in existing PV forecasting approaches, including single-horizon modeling, temporal-scale imbalance, and lack of interpretability.

The framework was evaluated on multiple real-world PV datasets from the Desert Knowledge Australia Solar Centre (DKASC), covering different years and climatic conditions. Experimental results demonstrated that the proposed model consistently outperformed traditional statistical models, machine learning methods, and state-of-the-art deep learning architectures. Specifically, the model achieved RMSE values of 42.35, 46.27, 52.29, and 48.11 across the evaluated datasets, representing a substantial performance improvement of approximately 35–40\% compared to the best-performing baseline (TFT). Additionally, MAPE values below 5\% across all datasets indicate high prediction accuracy and stability.

Beyond predictive performance, the proposed framework enhances interpretability through attention weight visualization and N-BEATS-based trend–seasonality decomposition, providing valuable insights into temporal dynamics and feature contributions. Furthermore, the deployment of the model on a cloud-based platform demonstrates its practical applicability for real-time forecasting in PV–battery–grid systems.

Future work will focus on extending the framework to incorporate probabilistic forecasting, integrating numerical weather prediction (NWP) data, and optimizing computational efficiency for large-scale deployment in smart grid environments.

\section*{Funding}
This research did not receive any specific grant from funding agencies in the public, commercial, or not-for-profit sectors.

\printcredits

\bibliographystyle{cas-model2-names}

\bibliography{cas-refs}



\end{document}